\documentclass{article}
\usepackage{spconf,amsmath,epsfig}

\usepackage{enumitem}
\usepackage{caption}
\usepackage{comment}
\usepackage{xcolor}
\usepackage{amsfonts}
\usepackage{ragged2e}
 \usepackage{amsmath}
 \usepackage{multirow}
\definecolor{bg}{rgb}{0.95,0.95,0.95}

\newcommand{\etcno}{\textit{etc}}
\newcommand{\egno}{\textit{e}.\textit{g}.} 
\newcommand{\ieno}{\textit{i}.\textit{e}.}

\let\OLDthebibliography\thebibliography
\renewcommand\thebibliography[1]{
  \OLDthebibliography{#1}
  \setlength{\parskip}{0pt}
  \setlength{\itemsep}{0pt plus 0.3ex}
}

\begin{document}\sloppy

\def\x{{\mathbf x}}
\def\L{{\cal L}}


\title{ADAPTIVE HIGH-FREQUENCY PREPROCESSING FOR VIDEO CODING}

%
\name{Yingxue Pang\sthanks{\{pangyingxue,zhaoshijie.0526,lijunlin.li,lizhang.idm\}@bytedance.com}, Shijie Zhao\sthanks{Corresponding author}, Junlin Li, Li Zhang}
\address{Bytedance Inc.}

  

\maketitle

\begin{abstract}
High-frequency components are crucial for maintaining video clarity and realism, but they also significantly impact coding bitrate, resulting in increased bandwidth and storage costs. This paper presents an end-to-end learning-based framework for adaptive high-frequency preprocessing to enhance subjective quality and save bitrate in video coding. The framework employs the Frequency-attentive Feature pyramid Prediction Network (FFPN) to predict the optimal high-frequency preprocessing strategy, guiding subsequent filtering operators to achieve the optimal tradeoff between bitrate and quality after compression. For training FFPN, we pseudo-label each training video with the optimal strategy, determined by comparing the rate-distortion (RD) performance across different preprocessing types and strengths. Distortion is measured using the latest quality assessment metric. Comprehensive evaluations on multiple datasets demonstrate the visually appealing enhancement capabilities and bitrate savings achieved by our framework.
\end{abstract}

\begin{keywords}
High-frequency preprocessing, bitrate and quality tradeoff, video compression
\end{keywords}
\section{Introduction}
\label{sec:intro}
Preprocessing is a crucial step in video coding as it enhances the quality and coding efficiency of compressed videos. Targeting high-frequency components, which often contain vital details, textures, and contours contributing to clarity and realism, is the most common and effective method. In this process, sharpening preprocessing is utilized to enhance high-frequency information for improved visual impact and sharpness. Simultaneously, smoothing preprocessing reduces high-frequency information to diminish noise, artifacts, and overshoot halos, thereby improving perceptual quality. Nevertheless, high-frequency components significantly impact the coding bitrate, necessitating higher bits for the accurate representation and reconstruction involved in coding and transmission. This leads to increased bandwidth and storage costs. Therefore, determining the appropriate high-frequency preprocessing strategy before encoding, tailored to different video content, is essential for achieving an optimal tradeoff between visual quality and bitrate.
\begin{figure}[!t]
\centering
\includegraphics[width=0.48\textwidth]{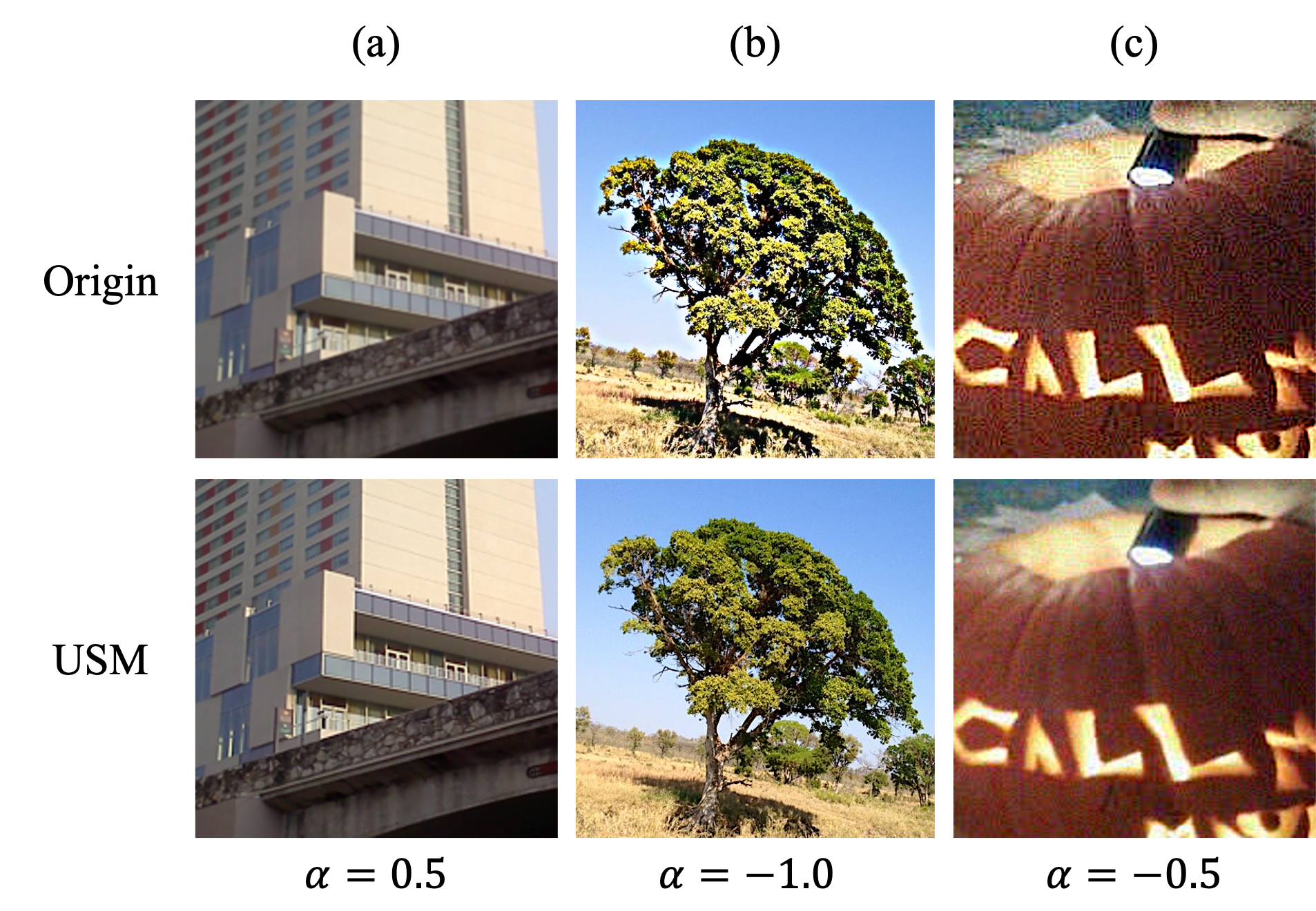}
\caption{Adaptive high-frequency preprocessing: (a) under-enhancement requires sharpening; (b) over-sharpening necessitates smoothing for perceptual improvement; (c) undesired noise or artifact also calls for smoothing.}
\label{fig:moti}
\end{figure}

This paper presents an end-to-end learning-based framework for high-frequency preprocessing in video coding, aiming to improve subjective quality and reduce bitrate. Our systematic framework primarily includes a \textbf{F}requency-attentive \textbf{F}eature pyramid \textbf{P}rediction \textbf{N}etwork (FFPN). FFPN adaptively predicts the optimal high-frequency preprocessing strategy for various content videos, guiding the preprocessing type and strength of subsequent filtering operators. Our framework provides the flexibility to apply sharpening or smoothing to high-frequency components, thereby maximizing perceptual quality improvement while minimizing required bits after compression.


To determine the optimal strategy for each training video, we pseudo-label them by comparing the rate-distortion (RD) characteristics of different high-frequency preprocessing strategies. Distortion is measured using the latest no-reference quality assessment metric, CLIP-IQA~\cite{clipiqa}. In addition, we directly adopt the widely used and computationally simple Unsharp Masking (USM) as the filter for high-frequency components preprocessing. We align it with our predicted optimal preprocessing strategy by adjusting its effect strength parameter $\alpha$. However, it's important to mention that other filters with similar functions, like Laplace filters or adaptive bilateral filters, are also valid options.


Mathematically, the preprocessed result $x_{usm}$ can be acquired by:
\begin{equation}
    x_{usm} = \alpha (x - \mathcal{F}_{L}(x)) + x.
    \label{eq:usm}
\end{equation}
where $x$ denotes the original image and $\mathcal{F}_{L}$ denotes the low-pass filter.
Fig.~\ref{fig:moti} illustrates the enhancement effects of adaptive high-frequency preprocessing on different types of distortion sources. In Fig.\ref{fig:moti}~(a), a positive $\alpha>0$ is applied as a sharpening filter, enhancing visual clarity by overlaying high-frequency components like edges and textures onto the original image. Fig.\ref{fig:moti}~(b) and (c) demonstrate the application of smoothing filters with different strengths. Negative $\alpha<0$ effectively reduces overshooting halos and artifacts in high-frequency components, leading to improved perceptual quality. The contributions of this paper can be summarized as follows:
\begin{itemize}
        \item We present a systematic framework that utilizes FFPN to predict the optimal strategy. This strategy informs the type and strength of preprocessing, guiding subsequent filtering operations efficiently and reliably. Our preprocessing framework achieves maximum perceptual quality gain with minimum bits after compression.

	\item We determine the optimal strategy for each training video via pseudo-labeling, which is done by comparing the RD performance of various high-frequency preprocessing types and strengths. We utilize the latest no-reference quality assessment metric to measure distortion.

	\item Comprehensive evaluation of our approach using multiple datasets, demonstrating its superiority through both quantitative and qualitative analyses.
\end{itemize}

\vspace{-2mm}
\section{Related Work}
\vspace{-2mm}

FreqSP~\cite{freqsp} recently proposed predicting the optimal sharpening level for balancing quality and bitrate tradeoff, primarily by comparing Bjøntegaard-Delta bitrate (BD-Rate) across different sharpening levels. However, this approach exclusively focuses on sharpening, neglecting the indiscriminate enhancement of high-frequency information, leading to amplified noise, artifacts and overshoots, particularly in user-generated content (UGC) videos with complex and uncertain degradation. We propose an adaptive high-frequency preprocessing framework including sharpening and smoothing, making it better suited for real-world videos with diverse distortions.

Furthermore, obtaining BD-Rate involves calculating the average quality difference between two RD curves across a bitrate range, which can be cumbersome. In practical scenarios, videos are transcoded to stream at specific candidate profiles or target bitrates to meet industrial requirements, making the comparison of RD performance over a continuous bitrate range (\ieno, BD-Rate) unnecessary. Our method ensures optimal RD at the target bitrate to determine the best strategy with type and strength, offering a more direct and precise approach.




\begin{figure}[!t]
\centering
\includegraphics[width=0.5\textwidth]{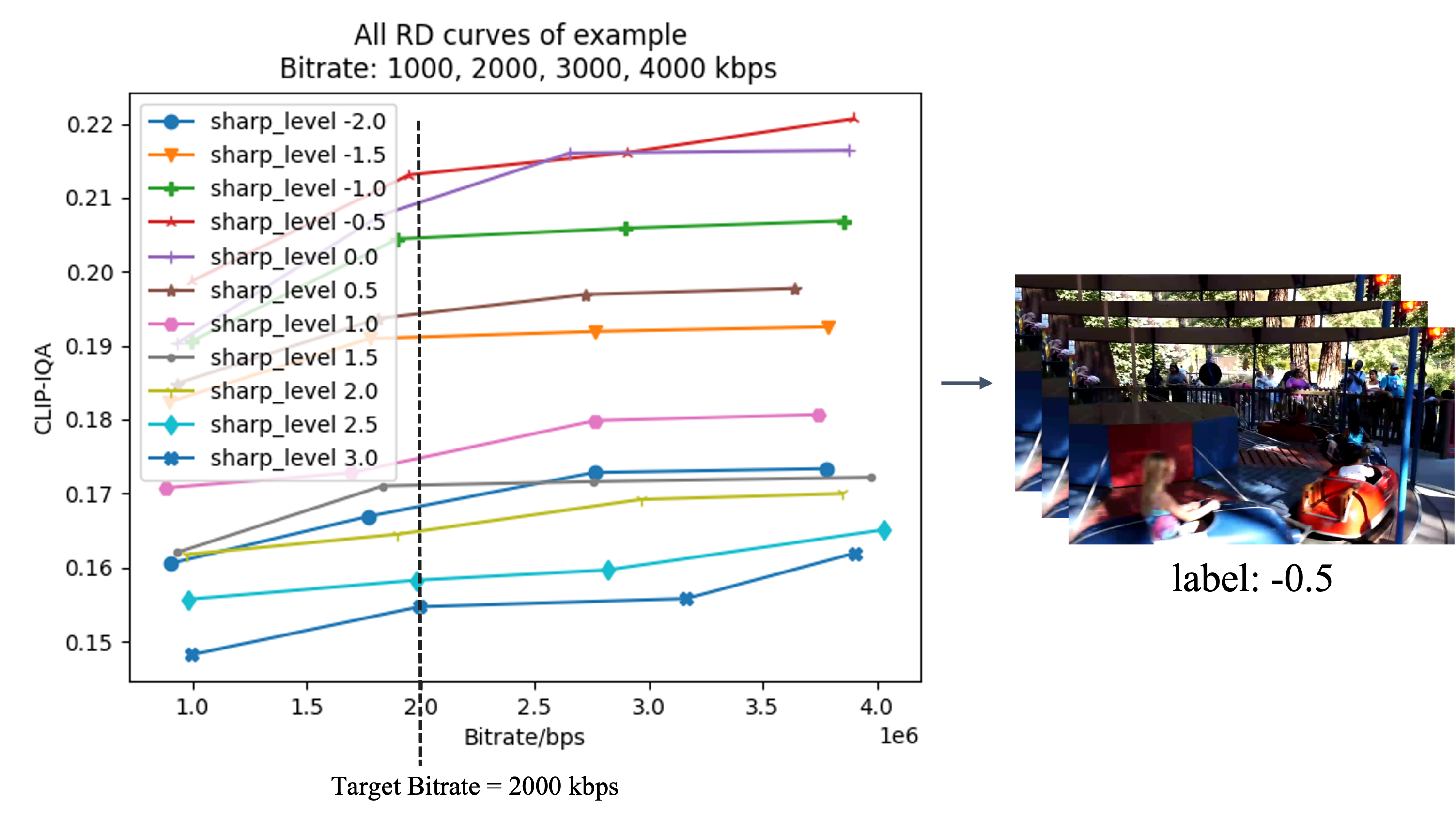}
\caption{Illustration of the pseudo-labeling process with RD.}
\label{fig:rd}
\end{figure}

\begin{figure}[!t]
\centering
\includegraphics[width=0.48\textwidth]{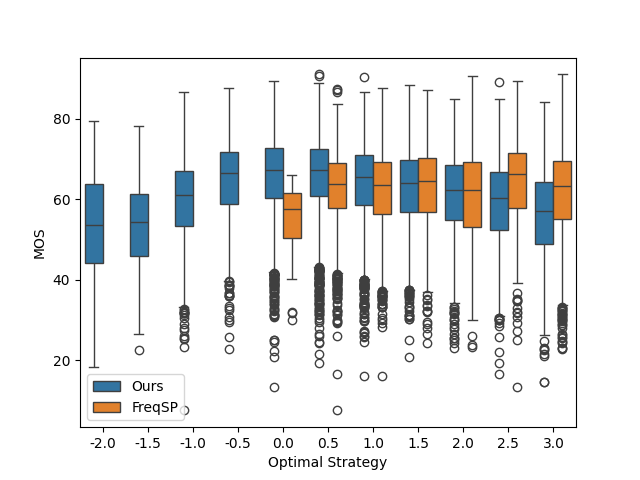}
\caption{Correlation between assigned pseudo-labels and MOS scores.}
\label{fig:corr}
\end{figure}

\begin{figure*}[!t]
\centering
\includegraphics[width=\linewidth]{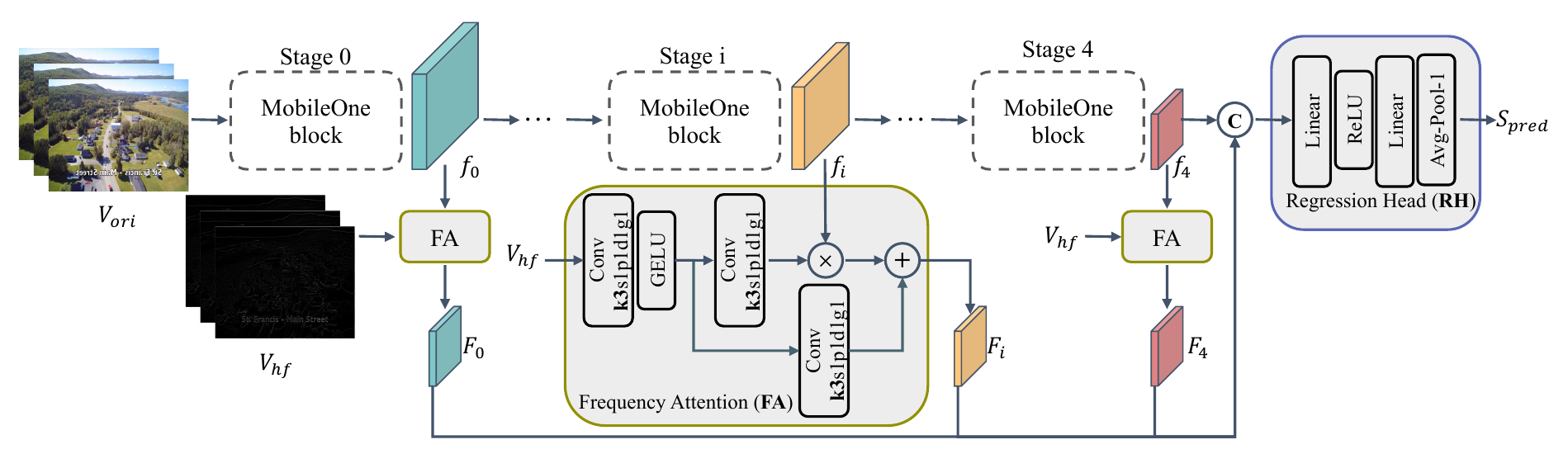}
\caption{Overall framework of our proposed FFPN.}
\label{fig:framework}
\end{figure*}

\vspace{-2mm}
\section{Methodology} 
\vspace{-2mm}
We propose a novel frequency-attentive feature pyramid prediction network to predict the optimal strategy for video high-frequency preprocessing before coding. This strategy informs the type (sharpening or smoothing) and strength of preprocessing to mitigate issues like under-enhancement, over-sharpening, and undesired noise or artifacts in source videos, while also saving bitrate after compression. In this section, we first elaborate on the process of pseudo-labeling to obtain the strength as ground truth for training, and then provide a detailed illustration of the careful architectural design of the proposed prediction model.

\vspace{-1mm}
\subsection{Pseudo Labeling}
\vspace{-1mm}
In video coding, RD theory~\cite{rd} evaluates encoder performance, with superior performance indicated by higher quality (\egno, PSNR, SSIM, VMAF, \etcno) at a specific bitrate. We use RD to determine the optimal strategy, selecting it directly for maximum perceptual quality gain at discrete target bitrates and achieving the best tradeoff between quality and bitrate in preprocessed compressed videos. For practical application, we employ a computationally simple USM filter for preprocessing high-frequency components. We equal its effect strength parameter ($\alpha$) to our ideal strategy, where positive and negative signs indicate the type of preprocessing (sharpening or smoothing), and numerical values indicate the intensity.

While widely-used metrics such as PSNR and SSIM optimize coding decisions and assess compressed video quality, they may not consistently align with human perception~\cite{vmaf,vmafbetter}. Although VMAF~\cite{vmaf} exhibits a better correlation with human opinion scores, it is susceptible to sharpening algorithms that inflate scores without enhancing actual quality~\cite{hackvmaf0,hackvmaf}. We propose using CLIP-IQA~\cite{clipiqa}, a no-reference metric designed to assess perceptual quality. Trained on the rich visual language embedded in contrastive language-image pre-training models (CLIP)~\cite{clip}, CLIP-IQA captures not only overall quality but also fine-grained dimensions such as brightness, noise, colorfulness, sharpness, naturalness, attractiveness, and aesthetics, offering a comprehensive evaluation of video quality. To assess a video with CLIP-IQA, we extract one frame per second (fps=1), compute the score for each frame, and calculate the average score as the final video quality assessment.


Fig. \ref{fig:rd} illustrates our labeling process, where a randomly sampled video from LSVQ \cite{lsvq} undergoes eleven strategies (-2.0 to 3.0) using the FFmpeg built-in USM function. The preprocessed videos are then compressed with HEVC/H.265 codec \cite{hevc} at four constant bit rate (CBR) settings (1000 kbps to 4000 kbps), and different encoders are denoted by the preprocessing compression process. By plotting the RD curves for each encoder with bitrate on the x-axis and CLIP-IQA on the y-axis, we identify the optimal strategy of $-0.5$ at the target bitrate of $2000$ kbps, which yields the maximum perceptual quality gain. Therefore, we assign a pseudo-label of $-0.5$ to the uncompressed source video for training.

Fig.~\ref{fig:corr} analyzes the correlation between the assigned pseudo-labels and the mean opinion score (MOS) for the original uncompressed video, providing additional evidence of our method's effectiveness. Our findings reveal that videos assigned the strategy of $0.0$ exhibit the highest MOS scores, reflecting an intuitive inclination to avoid or minimize preprocessing in high-quality raw videos. However, the sharpening level obtained through FreqSP~\cite{freqsp} using BD-Rate does not align with this pattern. Unfortunately, a video assigned the sharpening level of $2.5$ attains the highest MOS score, deviating from the expected results.

\vspace{-2mm}
\subsection{Overall framework}
\vspace{-2mm}

The proposed FFPN, as shown in Fig.~\ref{fig:framework}, comprises three key modules: pyramid feature extraction, frequency attention (FA), and feature regression head (RH). Each convolutional layer is denoted by its corresponding kernel size (k), stride (s), padding size (p), dilation (d), and groups (g). Hierarchical quality-aware features are extracted and fused with high-frequency components using an attention mechanism. These features are then mapped to optimal strategy through a regression head.

\begin{table}
\centering
\caption{Ablation studies on the prediction performance with different backbones on the LSVQ\cite{lsvq}, KoNViD-1k\cite{konvid1k} and LIVE-VQC\cite{livevqc} datasets.}
\label{table:cmp}
\resizebox{0.5\textwidth}{!}{  
\begin{tabular}{|c|cc|cc|cc|}
\hline
& \multicolumn{2}{c|}{$\text{LSVQ}_{test}$ (4028)} & \multicolumn{2}{c|}{KoNViD-1k (1164)} & \multicolumn{2}{c|}{LIVE-VQC (585)} \\ \cline{2-7} 
\multirow{-2}{*}{Backbone} & PLCC ($\uparrow$)   & RMSE ($\downarrow$)  & PLCC ($\uparrow$)   & RMSE ($\downarrow$)  & PLCC ($\uparrow$)      & RMSE ($\downarrow$)     \\ \hline
Resnet-18~\cite{resnet}               & 0.7905          & 0.5088          & 0.7964           & 0.6210           & 0.8039          & 0.4587          \\
Resnet-34~\cite{resnet}                & 0.7873          & 0.5127          & 0.8087           & 0.6003           & 0.8474          & 0.4046          \\
Repvgg-a0~\cite{repvgg}                & 0.7377          & 0.5718          & 0.7549           & 0.6773           & 0.8228          & 0.4356          \\
Repvgg-b0~\cite{repvgg}                 & 0.7320          & 0.5783          & 0.7482           & 0.6862           & 0.8226          & 0.4358          \\
FreqSP-11~\cite{freqsp}                  & 0.7775          & 0.5246          & 0.7835           & 0.6388           & 0.8290          & 0.4279          \\
FreqSP-15~\cite{freqsp}                  & 0.7903          & 0.5089          & 0.8073           & 0.6059           & 0.8374          & 0.4173          \\
\textbf{Ours} & \textbf{0.8189} & \textbf{0.4731} & \textbf{0.8132}  & \textbf{0.5975}  & \textbf{0.8491} & \textbf{0.4024} \\ \hline
\end{tabular}}
\end{table}

\begin{table}
\centering
\caption{Speed test of different backbones on GPU (A100-SXM-80GB) and CPU (Intel-Xeon-Platinum-8336C-CPU). The results of time usage are averaged over 20 runs after warming up the hardware with a single thread, using a 1080p video with 32 frames as input.}
\label{table:speed}
\resizebox{0.5\textwidth}{!}{ 
\begin{tabular}{|c|c|c|c|cc|}
\hline
\multirow{2}{*}{Backbone} & \multirow{2}{*}{Params/M} & \multirow{2}{*}{Memory/M} & \multirow{2}{*}{FLOPs/G} & \multicolumn{2}{c|}{Runtime/ms} \\ \cline{5-6} &     &               &    & $cpu_{t1}$           & $gpu_{t1}$       \\ \hline
Resnet-18~\cite{resnet}     & 12.23         & 618.84          & 152.57         & 3960.14         & 20.26         \\
Resnet-34~\cite{resnet}     & 22.34         & 705.00          & 307.61         & 13060.18        & 17.55         \\
Repvgg-a0~\cite{repvgg}     & 9.51          & 576.54          & 126.47         & 9309.62        & 16.80         \\
Repvgg-b0~\cite{repvgg}     & 16.34         & 737.52          & 284.04         & 38270.31       & 22.73         \\
FreqSP-11~\cite{freqsp}     & 2.85          & 1135.19         & 49.90          & 2146.53          & 10.42         \\
FreqSP-15~\cite{freqsp}     & 6.18          & 1170.13         & 86.00          & 3604.88        & 11.85         \\
\textbf{Ours} & \textbf{2.60} & \textbf{603.42} & \textbf{23.05} & \textbf{1249.13} & \textbf{6.46} \\
\hline
\end{tabular}
}
\end{table}

\textbf{Pyramid Feature Extraction} 
We expect to extract the quality-aware and bitrate-aware features that can represent the impact of subjective perception and compression effects using the predicted strategy. Human perception is significantly impacted by compression~\cite{perception1}, necessitating extraction of sufficient effective features from the original video for modeling and prediction. This encompasses both high-level semantic information and low-level details such as color, texture, shape, and brightness. High-level information aids in identifying perceptual distortions, while low-level information is crucial for preserving visual quality during compression. As such, we propose to extract multi-scale features from the four stages of the pre-trained MobileOne~\cite{mobileone} and utilize hierarchical features for regression prediction, progressing from low-level to high-level features.

\textbf{Frequency Attention}
We derive a high-frequency mask $V_{hf}\in{V_{ori}}$ from each input video $V_{ori}$ while preserving its spatial dimensions. To achieve this, we employ a 2D low-pass Gaussian kernel $\mathcal{F}_{L}$ to obtain the low-frequency information and subtract to filter the high-frequency feature. The process is as follows:
\vspace{-1mm}
\begin{equation}
    V_{hf} = Gray(V_{ori}) - \mathcal{F}_{L}(Gray(V_{ori})),
\end{equation}
\vspace{-1mm}
where Gray denotes the conversion of the color image to grayscale. This operation can avoid excessive enhancement of color information and typically we only process the luminance channel as it primarily contains the high-frequency structural and detailed information. We then propose frequency attention (FA) to progressively modulate the extracted hierarchical representations $f_{i}$ by utilizing a spatially adaptive, learned affine transformation with the generated high-frequency mask $V_{hf}$:
\vspace{-1mm}
\begin{equation}
    F_{i} = FA(V_{hf}, f_{i}).
\end{equation}
\vspace{-1mm}
The FA, shown in Fig.~\ref{fig:framework}, facilitates the propagation of high-frequency information throughout the network, emphasizing the significance of locally masked feature regions through spatially varying transformations.



\textbf{Regression Head}
Following the pyramid feature extraction and attentive frequency fusion, we concatenate and regress the fused multi-scale features using two-stage fully connected layers to obtain prediction values $S_{pred}$ via average pooling. The L1 loss $\mathcal{L}$ is employed:
\begin{equation}
    \mathcal{L} = |S_{pred}-S_{gt}|,
\end{equation}
where $S_{gt}$ denotes the ground truth.


\vspace{-2mm}
\section{Experiments}
\vspace{-2mm}

In this section, we present the experimental setup details and compare our proposed high-frequency prepressing strategy prediction model, FFPN, with the existing approach called FreqSP~\cite{freqsp}. Due to the limited related work in our proposed problem, we conduct a series of comparative experiments and ablation studies to thoroughly discuss the rationale behind each design choice in our model.

\begin{table}
\centering
\caption{Ablation studies on the number of MobileOne blocks and Frequency Attention (FA).}
\label{table:cmp2}
\resizebox{0.5\textwidth}{!}{  
\begin{tabular}{|c|cc|cc|cc|}
\hline
& \multicolumn{2}{c|}{$\text{LSVQ}_{test}$ (4028)} & \multicolumn{2}{c|}{KoNViD-1k (1164)} & \multicolumn{2}{c|}{LIVE-VQC (585)} \\ \cline{2-7} 
\multirow{-2}{*}{} & PLCC ($\uparrow$)   & RMSE ($\downarrow$)  & PLCC ($\uparrow$)   & RMSE ($\downarrow$)  & PLCC ($\uparrow$)      & RMSE ($\downarrow$)     \\ \hline
Stage 0,1         & 0.6778          & 0.8434          & 0.6719          & 0.8836          & 0.6708          & 0.8761          \\
Stage 0,1,2         & 0.7008          & 0.7924          & 0.6966          & 0.7979          & 0.7065          & 0.7904          \\
Stage 0,1,2,3        & 0.7887          & 0.5110          & 0.7994          & 0.6031          & 0.7852          & 0.5169          \\
\textit{w/o FA}        & 0.7374          & 0.5986          & 0.7640          & 0.6579          & 0.7539          & 0.5605          \\
\textbf{Ours} & \textbf{0.8189} & \textbf{0.4731} & \textbf{0.8132} & \textbf{0.5975} & \textbf{0.8491} & \textbf{0.4024} \\ \hline
\end{tabular}}
\end{table}

\vspace{-2mm}
\subsection{Experimental setups}
\vspace{-2mm}

We conduct experiments on three widely recognized natural video datasets, namely LSVQ \cite{lsvq}, KoNViD-1k \cite{konvid1k}, and LIVE-VQC \cite{livevqc}, which encompass diverse and realistic distortions. LSVQ and KoNViD-1k contain 20,142 and 1,164 in-the-wild UGC videos collected from web platforms, while LIVE-VQC comprises 585 videos captured by real cameras. In detail, we randomly select $80\%$ and $20\%$ of the videos from LSVQ for training and testing, respectively. We evaluate the superiority and generalizability of the proposed FFPN in a challenging cross-dataset setting on KoNViD-1k and LIVE-VQC. We utilize the Pearson linear correlation coefficient (PLCC) and root mean square error (RMSE) as prediction evaluation metrics, where a higher PLCC or lower RMSE indicates a more accurate numerical fit with the ground truth.

During training, we employ a commonly used video segmentation sampling method~\cite{tsn} for temporal sampling. The input video is partitioned into $16$ non-overlapping time segments. Within each segment, two consecutive adjacent frames are sampled at random locations, resulting in a 32-frame network input denoted as $V_{ori}$. For spatial sampling, each frame is divided into $16\times16$ uniform grids with the same sizes. From each grid, we extract $16\times16$-sized patches at random positions. These patches are then recombined based on their respective grid positions and order, forming a $256\times256$-sized image that serves as the input to the network.

We utilize the MobileOne~\cite{mobileone} as the FFPN backbone for pyramid feature extraction, pre-trained on the ImageNet with MobileOne-S0 weights. We employ the AdamW optimizer with an initial learning rate of 
$0.001$ for the MobileOne backbone and $0.01$ for FA and RH, respectively. If the training loss does not decrease for $5$ epochs, the learning rate is halved. The default number of epochs and batch sizes are 
$30$ and $16$, respectively. Our model is implemented using the PyTorch framework on a single NVIDIA A100-SXM-80GB GPU.


\begin{figure}[!t]
\centering
\includegraphics[width=0.5\textwidth]{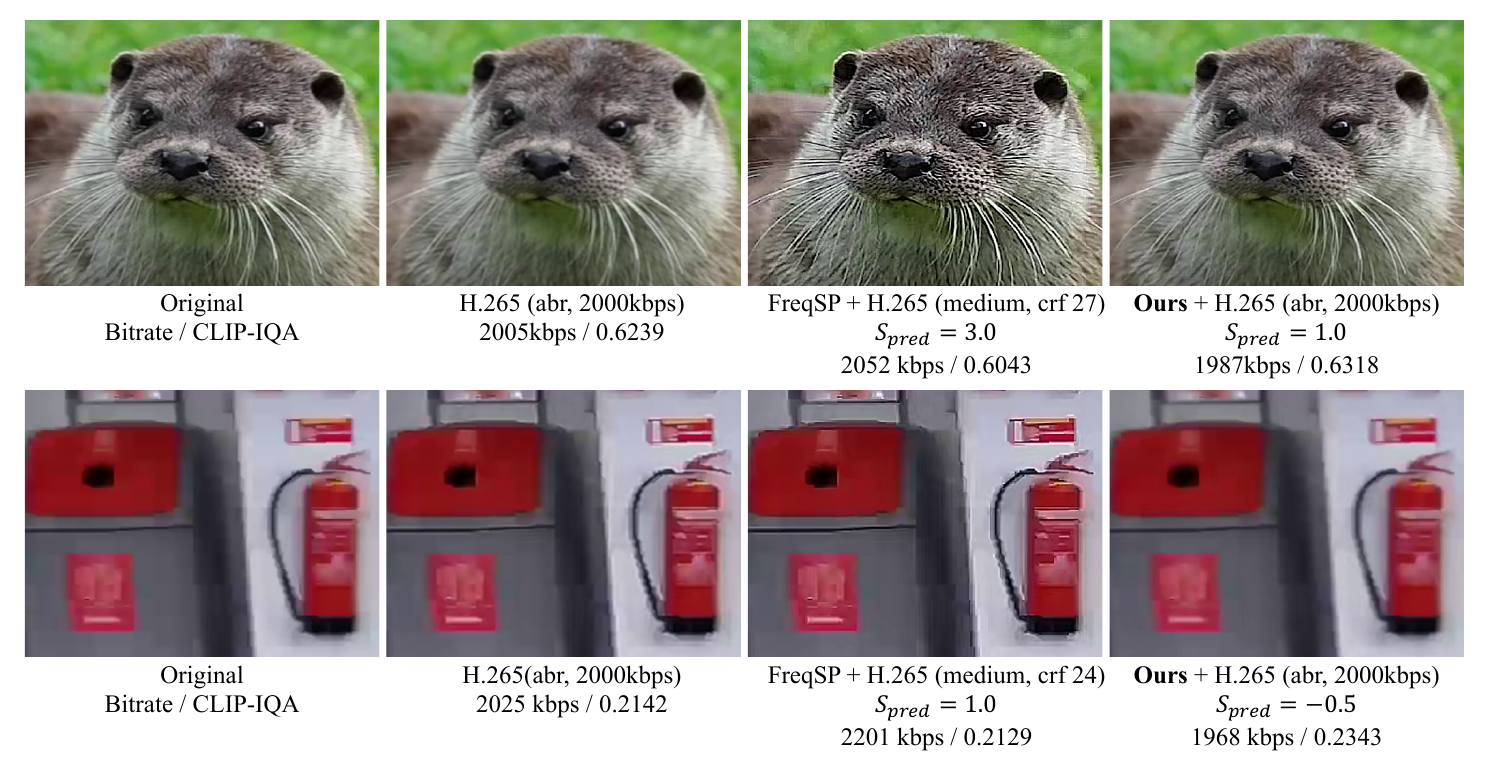}
\caption{The subjective quality comparison of enhancement after compression. (\textbf{Zoom-in for best view})}
\label{fig:cmp}
\end{figure}

\begin{figure}[!t]
\centering
\includegraphics[width=0.4\textwidth]{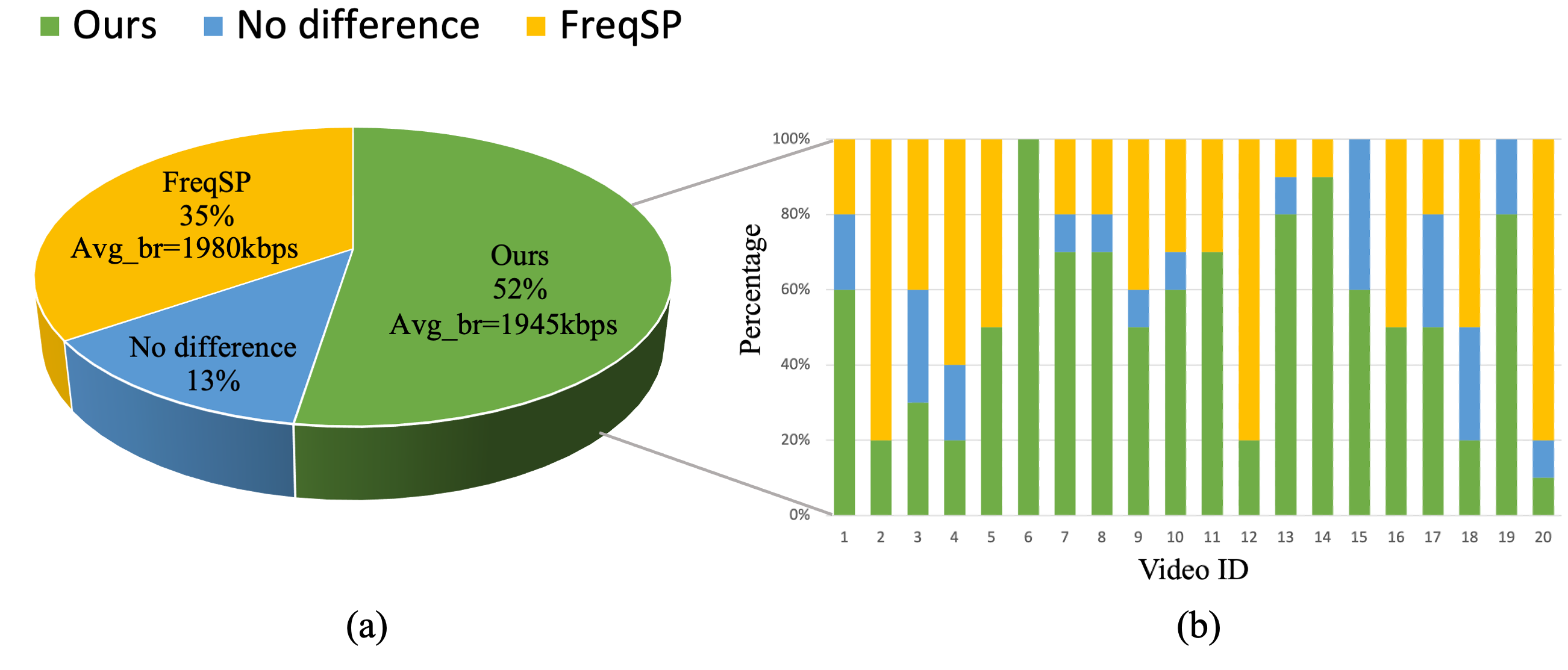}
\caption{User study results: (a) overall voting percentage and average bitrate (Avg br); (b) voting percentage for each video.}
\label{fig:vote}
\end{figure}
\vspace{-4mm}

\subsection{Results and Analysis}
The competing model, FreqSP, is re-trained on our pseudo labels using the provided training codes from the respective authors~\cite{freqsp}. The choice of backbone significantly impacts prediction performance. We compare our proposed FFPN with four other backbones: ResNet-18~\cite{resnet}, ResNet-34~\cite{resnet}, RepVgg-a0~\cite{repvgg}, and RepVgg-b0~\cite{repvgg}. As depicted in Table~\ref{table:cmp}, our FFPN demonstrates remarkable performance on all three datasets (mean PLCC $>0.82$). It outperforms FreqSP-15 with an average PLCC improvement of $2.88\%$ and an average RMSE drop of $-0.02$. Comparatively, the four alternative backbones perform notably worse, while our FFPN achieves state-of-the-art results in strategy prediction. This further indicates the superior efficiency of our model in extracting pyramidal features related to bitrate and perceptual quality, enabling better modeling and accurate strength prediction. 


We also evaluate the efficiency of various backbones in terms of parameter count (Params/M), memory consumption (Memory/M), computational complexity (FLOPs), and the running time on both Intel-Xeon-Platinum-8336C-CPU ($cpu_{t1}$) and A100-SXM-80GB-GPU ($gpu_{t1}$). The results of time usage are averaged over 20 runs after warming up the hardware with a single thread, using a 1080p video with 32 frames as input. Table~\ref{table:speed} shows the superior performance of our model across all efficiency metrics, enabling faster inference and requiring fewer resources, making it suitable for resource-constrained environments and industrial deployment.

To verify our architecture design, we investigate the impact of the number of MobileOne blocks for pyramid feature extraction and the frequency attention (FA) module. Table~\ref{table:cmp2} demonstrates that increasing the number of MobileOne blocks leads to better performance. Among them, our model (Ours) reaches the best performance by fully leveraging the advantages of the feature pyramid and incorporating enriched information across all scales. Additionally, we examine the effect of FA on the accuracy of strategy prediction by comparing it against a variant without FA (\textit{w/o FA}). The inclusion of FA, in conjunction with extracted hierarchical features, results in approximately $10\%$ improvement in PLCC and $-0.11$ decrease in RMSE across three datasets.

To evaluate the RD performance of the predicted strategy $S_{pred}$, we utilize it for USM filtering and compress the preprocessed video using H.265's CBR mode at $2000$ kbps. FreqSP utilizes the original version provided by the authors ~\cite{freqsp}, instead of our retraining one, and employs CRF coding for H.265 compression, selecting the bitrate closest to $2000$ kbps. Fig.~\ref{fig:cmp} shows that our results enhance clarity without excessive sharpening, reduce compression artifacts and edge jaggedness, and achieve bitrate savings compared to FreqSP. We further conduct a user study to compare the subjective effects of our method and FreqSP. We randomly select 20 videos from the testing data. Each video is displayed with three results on a webpage: A (H.265), B (Ours + USM+ H.265), and C ( FreqSP + USM + H.265). Participants ($n=20$) are asked to select their preferred result between B and C (in random order), or indicate no difference. We collect a total of 400 responses (20 videos multiplied by 20 participants). Fig.~\ref{fig:vote} shows a higher preference for our method with lower bitrate.

\vspace{-1mm}
\section{Conclusion}
\vspace{-1mm}
In this paper, we present an adaptive preprocessing framework for video coding, aiming to enhance the subjective quality and reduce bitrate consumption of high-frequency information. Our framework utilizes FFPN to predict the optimal preprocessing strategy for source videos with varying qualities and guides subsequent filtering operations by employing flexible preprocessing types and strengths. We pseudo-label the optimal strategy for each training video by comparing the RD performance of different preprocessing strategies. Through both quantitative and qualitative evaluations, our approach demonstrates its superiority.

\bibliographystyle{IEEEbib}
\bibliography{main}

\end{document}